\documentclass{article}

\usepackage{microtype}
\usepackage{graphicx}
\usepackage{subcaption}
\usepackage{booktabs} 

\usepackage{hyperref}

\usepackage[preprint]{icml2026}

\usepackage{amsmath}
\usepackage{amssymb}
\usepackage{mathtools}
\usepackage{amsthm}

\usepackage[capitalize,noabbrev]{cleveref}

\theoremstyle{plain}

\theoremstyle{definition}

\theoremstyle{remark}

\usepackage[textsize=tiny]{todonotes}

\icmltitlerunning{Interpretable experiential learning based on state history and global feedback}

\begin{document}

\twocolumn[

  \centering{\textit{This work has been submitted to the IEEE for possible publication. Copyright may be transferred without notice, after which this version may no longer be accessible.}}

  \icmltitle{Interpretable experiential learning based on state history and global feedback}



  \icmlsetsymbol{equal}{*}

  \begin{icmlauthorlist}
    \icmlauthor{Anton Kolonin}{univer,comp}
  \end{icmlauthorlist}

  \icmlaffiliation{univer}{The Artificial Intelligence Research Center, Novosibirsk State University, Russia}
  \icmlaffiliation{comp}{Aigents, Russia}
  
  \icmlcorrespondingauthor{Anton Kolonin}{akolonin@gmail.com}

  \icmlkeywords{edge computing, experiential learning, explainable AI, global feedback, interpretable AI, low-end computing, reinforcement learning, psyche model, resource-constrained computing, state history}

  \vskip 0.3in
]



\printAffiliationsAndNotice{}  

\begin{abstract}
A new interpretable experiential learning model based on state history and global feedback is presented. It is capable of learning a behavioral model represented by a transition graph between sets of states, with transitions attributed with utility and evidence count. This model is expected to be suitable for solving reinforcement learning problem in resource-constrained environments. The model was thoroughly evaluated on the OpenAI Gym Atari Breakout benchmark, demonstrating performance comparable to some known neural network-based solutions.
\end{abstract}

\section{Introduction}
\label{introduction}

Reinforcement learning (RL) technology has undergone rapid development over the past 10 years \cite{oh2025,yang2025benchmarkingincontextexperientiallearning,machines13121140}. Significant advances have been made in virtual gaming environments, where deep reinforcement learning algorithms have demonstrated performance significantly superior to human capabilities \cite{mnih2013playingatarideepreinforcement,toromanoff2019deepreinforcementlearningreally,Schrittwieser_2020,badia2020agent57outperformingatarihuman}. 

However, the deep reinforcement learning-based methods used in the above-mentioned works are limited in terms of providing interpretability and trustworthiness with respect to the learned models, which remains problematic for the application of such methods in mission-critical industrial automation tasks \cite{10.1145/3527448,machines13121140,9619637} or even less critical home automation tasks. In \citet{10.1145/3527448} we find “Interpretability, explainability, and transparency are key issues to introducing artificial intelligence methods in many critical domains ... Reinforcement learning methods, and especially their deep versions, are such closed-box methods.” In particular, in respect to industrial automation, \citet{machines13121140} calls for “practical industrial applications ... emphasizing transparency, sustainability, and operational resilience.” 

Moreover, “smart home” automation applications \cite{10.1145/3642975.3678961,10724603,en17246420}, as well as other potential consumer electronics applications, can be considered examples of “edge computing,” “resource-constrained computing” or “low-end computing,” where computing power is limited and the use of GPU cards or even clusters of such cards for deep reinforcement learning is not feasible. \citet{Farooq_2024} states the “prevalent challenges encountered in RL optimization, including issues related to sample efficiency and scalability; safety and robustness; interpretability and trustworthiness.”  

An additional problem, typical for real automation scenarios, is the sparse, delayed or implicit reward, which forces us to rethink the concept of “reinforcement learning” as a marginal case of “experiential learning” \cite{10.1007/978-3-030-93758-4_12,yang2025benchmarkingincontextexperientiallearning,kolonin2025computationalconceptpsychein} where feedback from the environment may not necessarily be rewarding, and environmental observations can be used to infer implicit feedback, while explicit feedback may be applicable over long periods of observation, according to “global feedback” principle in \citet{10.1007/978-3-030-93758-4_12}.     

Another key challenge in this area is the need for dimensionality reduction in the form of representation learning \cite{10.1109/TPAMI.2013.50}, which reduces the high-dimensional space of the field's input data to a low-dimensional space of hidden or latent variables. In terms of the required interpretability \cite{10.1145/3527448,machines13121140,Farooq_2024}, we can consider this low-dimensional space to be interpretable, represented by reasonable features, objects, and events, as shown in \citet{10.1007/978-3-030-93758-4_12}.

In this study, we focus on solving the problem of sparse or delayed reward mentioned above under interpretability constraints. In this context, the proposed solution creates interpretable representations of domain-specific problems through pre-processing that transforms time series of states of the original field data into time series of interpretable objects, events, and properties associated with these objects. These interpretable states and their historical sequences then themselves become part of the inferred model without introducing additional hidden or latent states.

Since notable progress in reinforcement learning is being made using the OpenAI Gym Atari benchmarks \cite{10.5555/2566972.2566979}, we evaluate our solution in the context of the Atari game “Breakout”, referring to earlier experiments described in \citet{mnih2013playingatarideepreinforcement,10.5555/3454287.3455074,toromanoff2019deepreinforcementlearningreally,kapturowski2019recurrent,badia2020uplearningdirectedexploration,badia2020agent57outperformingatarihuman,Schrittwieser_2020,pivovarov2025marti5mathematicalmodelself}. The game “Breakout” was chosen as one of the most popular OpenAI Gym environments, offering an experience similar to a single-player version of a real “ping-pong” game.

\section{Related Work}
\label{related_work}

\subsection{OpenAI Gym Atari Breakout Baselines}

The use of OpenAI Gym Atari environments as a platform for developing reinforcement learning algorithms has become popular since \citet{10.5555/2566972.2566979}. In these environments, each game is modeled by a \textit{step} function, which takes an \textit{action} argument and returns a \textit{observation} structure representing the changed state of the environment corresponding to the next visual frame, a \textit{reward} value containing explicit feedback, \textit{terminated} and \textit{truncated} variables signaling the end of the game, and an \textit{info} structure providing some additional information that can be used to obtain implicit feedback, such as  game termination in case of the “Breakout”. Each interaction with the \textit{step} function corresponds to one frame of the actual video game if \textit{frameskip=1} is set. During a game, the player either earns points by hitting bricks on the opposite side of the game field, moving the racket (paddle) with appropriate actions, or loses a life. If all bricks are destroyed, or 864 points are scored, or 5 lives are lost, the game ends. The number of points achieved at the end of the game is considered the game score. In the case of the “Breakout,” the maximum number of steps or frames is 108,000. However, the scoring methodology used in \citet{10.5555/2566972.2566979} calls for the game to end after 18,000 frames (corresponding to 5 minutes of play) if it has not already been completed.

Quick progress in this area began with the work of \citet{mnih2013playingatarideepreinforcement}, which applied the deep Q-learning  (DQN) method. For the game “Breakout,” an average score of 168 with a maximum possible score of 225 was achieved after 50 training epochs (corresponding to 5,400,000 training frames), with each epoch consisting of 30 minutes of play (108,000 frames). This study also evaluated the average score of 31 achieved by an expert human game player. Another baseline was achieved by implementing an extended version of the DQN algorithm called Rainbow in combination with the Implicit Quantile Networks (IQN) method \citet{toromanoff2019deepreinforcementlearningreally}, which takes into account probability distributions on state transition graphs during training. The new algorithm achieved average scores of 53.77, 121.83, 132.56, and 175.47 for training frames in the millions of 10, 50, 100, and 200, respectively.

Progress accelerated with the implementation of distributed multitasking capacity and history-based experience replay in the Recurrent Replay Distributed DQN (R2D2) architecture, achieving 837-864 scores under different experimental conditions \cite{kapturowski2019recurrent}. The inclusion of implicit feedback (called intrinsic reward) was introduced in  Never-Give-Up (NGU) architecture, based on extrinsic (explicitly provided by the environment) and intrinsic (implicitly derived) rewards (feedback) based on exploratory activity
\citet{badia2020uplearningdirectedexploration}, achieving 576-864 scores across different evaluation conditions. Another multi-agent distributed RL architecture, Agent57, combined history-aware DQN, IQN, and NGU with exploratory capabilities and achieved a score of 790 \cite{badia2020agent57outperformingatarihuman}. A record score of 864 was achieved by the latest MuZero architecture, which added planning capabilities to other features \cite{Schrittwieser_2020}, where planning was based on a hidden model of the environment learned through interaction with it.

\subsection{Interpretable Approaches}

A review of earlier work on interpretable reinforcement learning approaches is presented in \citet{glanois2022surveyinterpretablereinforcementlearning}. The most recent paper \cite{10.1145/3587716.3587798} presented the possibility of learning so-called “Behavior Trees” and evaluated the results on several classic OpenAI Gym environments, confirming the effectiveness of the method. Furthermore, the paper \citet{paleja2023interpretablereinforcementlearningrobotics} presented the possibility of learning “Interpretable Continuous Control Trees” (ICCTs) and confirmed the practical application of the method in real field conditions.

In \citet{10.1007/978-3-030-93758-4_12,kolonin2025computationalconceptpsychein}, the feasibility of constructing models based on non-latent explicit representations of environmental states, either on high-dimensional raw field data (termed “discrete”) or on interpretable low-dimensional representations of the environment in terms of real-world objects and events (termed “functional”), was investigated. A mathematical model of a single-player ping-pong game with rules similar to those of the OpenAI Gym Atari “Breakout” was used, and it was confirmed that the same learning model, based on the history of states with so-called “global feedback”, can be used to achieve the same level of performance in the same environment using both of two representations under the same evaluation conditions. The only difference was that learning and execution based on high-dimensional “discrete” (raw field) data turned out to be much slower and computationally expensive than learning based on low-dimensional “functional” representation. “Global feedback” implied that the utility values of state transitions, corresponding to actions specific to each state, were updated not gradually, as in DQN, but rather over entire sequences of states in the agent's episodic memory, so that each state and action in the corresponding sequence was updated equally. The boundaries of the state sequences were determined by the moments of receiving either positive feedback (the ball hitting the paddle or the opposite wall) or negative feedback (the ball missing the paddle on the player's side).

\citet{pivovarov2025marti5mathematicalmodelself} approached the reinforcement learning problem from a neuro-physiological plausibility perspective by constructing a mathematical model simulating cortical columns for learning symbol-like representations that support interpretability and computational efficiency. The model was tested in the OpenAI Gym Atari “Pong” (a ping-pong game with an artificial opponent) and “Breakout” environments. In the latter environment, the maximum score was 420, and the average was 50 after 50,000 completed games (approximately 170 million steps).

\section{Computational Model and Architecture}
\label{approach_method}

The computational model used in our solution was based on the concept of a “vector model of psyche,” based on earlier works by \cite{Kotliarov2007,petrenko2012goal,kolonin2025computationalconceptpsychein}. According to our interpretation of this concept, an agent perceives vectors $s_t$ of observations from the external environment and internal processes at each moment in time $t$, where $s_t$ consists of sensor observations $f_t$, internal motivations $y_{t}$, and awareness of completed actions $a_{t}$.

The agent maintains a historical memory of such state vectors as a sequence of $T$ states and can construct a model of the world's interactions as a transition graph between such sequences $\{s_{\tau}\}_{\tau\in\{-T,-1\}} => s_{\tau=0}$, where $\tau$ is the relative time for a given transition subgraph. Each transition in the model can be assigned its utility $U$ for satisfying the agent's motivations or needs, as well as a transition probability $P$ or a raw evidence count $C$ that determines this probability.

The learning function $L$ can compute the incremental update of utility $U$ when transitioning from the previous state at time $t-1$ to the new state at time $t$ as follows, where $x$ is the vector of importance of current motivations or needs, $y_t-y_{t-1}$ is the satisfaction or dissatisfaction assessment, interpreted as a “reward” in the context of reinforcement learning, corresponding to positive or negative “feedback” in our work. $E(a_{t-1})$ can be estimated as the amount of energy lost in case the agent seeks to minimize resource losses, but this was not considered in our work. In this study, we considered $y$ as the desire to obtain scores satisfying the new metrics and the avoidance of “loss of life” with equal weights in $x$ (so we used $L(y_t-y_{t-1}) = y_t-y_{t-1}$).

$dU(\{s_{\tau}\}_{\tau\in\{-T,-1\}},s_{\tau=0}) = L(x\cdot(y_t-y_{t-1}),E(a_{t-1}))$

According to the “global feedback” principle proposed and evaluated in \citet{10.1007/978-3-030-93758-4_12}, the utility update applies to all transitions during the observation period that correspond to a significant historical episode or sequence of states between the moments of receiving positive (a brick is destroyed on the opposite side) or negative (losing a life during the game) feedback.

The decision making function can be based on maximizing the utility $U$ of transitioning from a given state sequence $\{s_{\tau}\}_{\tau\in\{-T,t-1\}}$ at time $t-1$ to a new state $s_t$ at time $t$, or on maximizing some function, in our study, the multiplication, combining the utility $U$ and the probability $P$ or the evidence count $C$. The choice was based on the “counted utility” (CU) hyper-parameter. That is, to select the next action, the most recent state sequence was searched in the model to determine the hypothetical transition point. The number of states to index during training and search was determined by the context size (CS) hyperparameter. If no state sequence identical to the current situation was found, the most similar state sequence could be found based on the cosine similarity between the actual and hypothetical state sequences exceeding the state similarity (SS) threshold hyperparameter. Then the expected next state $s'_t$ can be determined as follows, depending on CU policy. Once the expected state $s'_t$ is determined, the selected action vector $a'_t$ is extracted from it for execution.

$s'_t=\arg \max_{s} U(\{s_{\tau}\},s_{\tau=0}), \tau\in\{-T,-1\}$, CU=$False$

$s'_t=\arg \max_{s} U(\{s_{\tau}\},s_{\tau=0}) \cdot C(\{s_{\tau}\},s_{\tau=0})$ , CU=$True$

In our study, the computational architecture included three layers: a state transformer, a state learning layer, and a decision making layer. The reference implementation, experimental setup, and presented results are available at \url{https://github.com/aigents/aigents-python}.

The state transformer was designed to process the input data, which was a multidimensional grayscale representation of a pixel map, and transform it into an interpretable representation of objects and events specific to the environment. We used a custom simple computer vision algorithm that identified pixel regions on the game field that were distinct from the background, presumably associated with the ball and paddle images in the Atari Breakout game, and used only the horizontal coordinates of these pixel clouds as independent state variables, used as elements of the vector $f_t$ to form the state vector $s_t$. Other state elements included the feedback vector $y_t$ (positive and negative feedback variables), if the “state reward” (SR) hyperparameter was set, and the action vector $a_t$.

The state learning component was based on an in-memory graph database capable of indexing transitions from some sequences of states to another states, where each state, sequence of states, and transition can be attributed with a learned utility and an evidence count. During interaction with the environment, the entire history of states $s_t$ was accumulated. When positive or negative feedback was detected, the current history of states at the time of receiving the feedback was remembered in the graph, such that each transition from $\{s_{\tau}\}_{\tau\in\{-T,-1\}}$ to $s_{\tau=0}$, where $T$ corresponds to the context size (CS) hyperparameter, was memorized with a corresponding increase in the evidence count $C$ and an increase or decrease $dU$ in the utility $U$ in the case of positive or negative feedback, respectively, in accordance with the principle of “global feedback”.

The decision component received the transformed state $s_t$ and combined it with previous states to create a search key with the context size (CS), which allowed retrieving the corresponding entry from the transition graph based on either an exact match or the highest similarity, as discussed above. The decision to transition to a new state from among possible options was made based on $U$ or $U \cdot C$, based on the “counted utility” (CU) hyperparameter. If no suitable sequence of states or transitions to a new state was found to determine the  action, a random action was considered.

\subsection{Hyper-parameters}
\label{hyper-parameters}

\begin{itemize}
    \item Context size (CS): the number of states representing the current situation and indexed in the model in a state transition graph, where the sequence of states CS is considered as a key, and the next state is considered as a value to which the utility of the transition and its evidence count in the training history are assigned (default is CS=2)
    \item Learning mode (LM): 0 - no learning, 1 - learning on positive feedback only, 2 - learning on positive and negative feedback both (default is LM=2)
    \item State reward (SR): specifies if positive and negative rewards are included in the state ($True$, default) or not ($False$) 
    \item Counted utility (CU): specifies the utility of the state transition is weighted by its evidence count value learned in the model ($True$) or not ($False$, default) 
    \item Encode action (EA): specifies if the action is represented in the state as a one-hot encoded vector ($True$) or a scalar value ($False$, default)
    \item State count threshold (SC): specifies the minimum count of experiences of series of states (of size SC) to be considered as a reference for searching for possible transitions (default is SC=2)
    \item State similarity threshold (SS): the minimum similarity of a series of states (of CS size) that can be considered applicable to use as the most similar state when searching for a possible transition if no ideal candidate for the current situation (of CS size) is found (default is SS=0.9)
    \item Transition utility threshold (TU): indicates the minimum utility of a transition in learning history for it to be considered as a candidate (default is TU=0)
    \item Transition count threshold (TC): indicates the minimum evidence count of a transition in learning history for it to be considered as a candidate (default is TC=1)
\end{itemize}

\section{Experiments}
\label{experiments}

The experimental work in this study was carried out in three stages. In the first phase, we implemented a framework capable of “imitation learning” and assessed the ability of the framework to remember a history of states associated with positive or negative feedback, and then use these memories to act as if they had been learned through experience.

In the second phase, we conducted a cursory study of our framework's ability to learn from scratch without limit on the number of steps per game.

In the third phase, we performed a systematic study of the learnability of our framework for different values of the random seed and evaluated different combinations of hyper-parameters.

The OpenAI Gym environment for Breakout was initialized using the setup \textit{gym.make('BreakoutNoFrameskip-v4', obs\_type="grayscale")}, so that the frame skipping option was not set, and each frame was evaluated as a separate step, with a grayscale pixel array used for observations.

All experiments were conducted on low-cost computing equipment consisting of two laptops: 1) MSI Raider GE77HX 12UGS notebook with 12th Gen Intel(R) Core(TM) i7-12800HX 2.00 GHz, 32.0 GB RAM Laptop GPU running Windows; 2) MacBook Pro with 2.9 GHz 6-Core Intel Core i9, Radeon Pro 560X 4GB Intel UHD Graphics 630 1536 MB, 32 GB 2400 MHz DDR4 running MacOS. Python software version 3.11.13 was used.

The average performance during runtime evaluation was 100-200 steps or frames per second. Given 18,000 frames, corresponding to 5 minutes of game-play, and an average frame rate of 60 frames per second, real-time training during game-play could have been achieved. However, all evaluations were conducted with rendering disabled and no display output, so the speed of the experiments was 2-3 times higher than real-time. In reality, when running experiments in batches of 1,000 games, each batch took between 4 and 18 hours, depending on the number of actual steps in the game, with up to 7 instances of the evaluation environment running on a single laptop. The total time spent on the study, including computing resources, was approximately 6 weeks on both laptops.

\begin{figure*}[ht]
  \vskip 0.2in
  \begin{center}
    \centerline{\includegraphics[width=1.7\columnwidth]{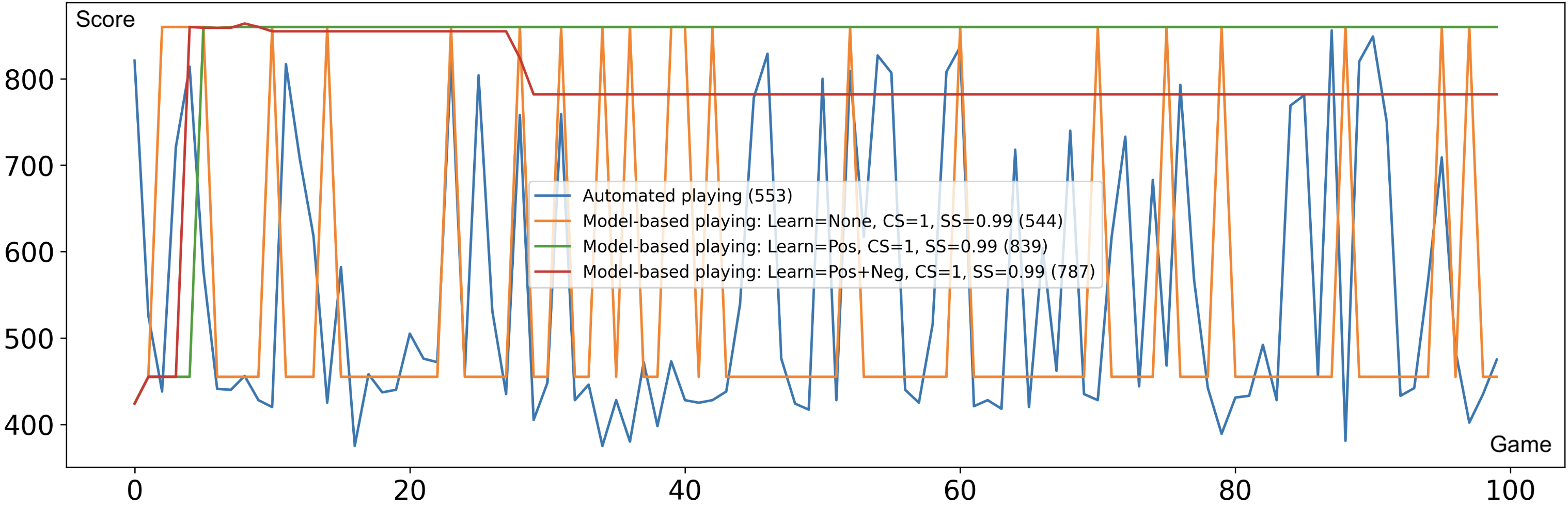}}
    \caption{
      Scores earned in four different runs playing 100 games. Horizontal axis - games from 1 to 100. Vertical axis - scores per game. Blue - “Automated” agent following the game rules based on pre-processed input providing tentative horizontal coordinates of the ball and the paddle. Orange - “Model-based” playing using the model pre-trained by “Automated” agent without the ability to learn. Green - “Model-based” playing using the same pre-trained model with learning based only on positive feedback. Red - “Model-based” playing using the same pre-trained model with learning based on both positive and negative feedback. Context size CS=1, state similarity threshold SS=0.99. Numbers in parentheses in the legend indicate average scores.
    }
    \label{fig:round1}
  \end{center}
\end{figure*}

\subsection{Phase 1: Exploring Decision Making Ability}
\label{sec:phase1}

We implemented an “Automated” agent that knew the rules of the Breakout game. The agent received the ball and paddle coordinates from the pre-processing layer and executed actions to hit the ball based on its location. The resulting states, including the previous action that could influence the transition to that state, were stored in a state transition memory, and the every transition utility values were updated based on the received feedback. The “Automated” agent was tested in 100 games, which were played with an average score of 553 and a maximum score of 856, as shown in \autoref{fig:round1} (the “Automated playing” is shown in blue). The imperfect results (below 864) can be explained by the known “right wall” issue in the OpenAI Gym Breakout game, where the actual behavior of the paddle during play violates the assumed physical laws, breaking the wall when moving to the right.

Following this “imitation learning” training phase, the “Model-based” agent was subjected to another 100 games using the model learned by the previous agent, but without the opportunity to improve it during the game (LM=0), so the collected feedback was not used to update its memory. In this case, using the pre-learned model improved the maximum score to 860 (possibly due to randomness) with a slight decrease in the average score to 544, as shown in \autoref{fig:round1} (“Model-based playing, Learn=None”).

The same agent using the same model was then run through 100 more games twice with different learning settings, as shown in \autoref{fig:round1}: “Model-based playing, Learn=Pos” to account for only positive feedback (LM=1); “Model-based playing, Learn=Pos+Neg” to account for both positive and negative feedback (LM=2). In these runs, the agent was able to learn as it played, gradually improving the model as it played. With only positive feedback, the average score was 839, and the maximum score was 839. With both positive and negative feedback, the average score was 787, and the maximum score was the maximum possible score of 864.

During the experiments described above, the context size was set to CS=1, so that only one state from history was used to search for a suitable situation and consider alternative transitions to new states. The state similarity threshold for searching for approximately suitable situations was set to SS=0.99. The initial value of the random number generator was not controlled, so completely random values based on the system time were used in all runs.

This experiment confirmed the framework's ability to use a pre-learned (pre-trained) model to effectively perform actions (inference) with the ability to quickly improve it as it runs. Rapid performance improvements were observed in both training cases (with and without negative feedback) within the first ten games.

\subsection{Phase 2: Exploring Learning Ability - Cursory}

\begin{figure*}[ht]
  \vskip 0.2in
  \begin{center}
    \centerline{\includegraphics[width=1.7\columnwidth]{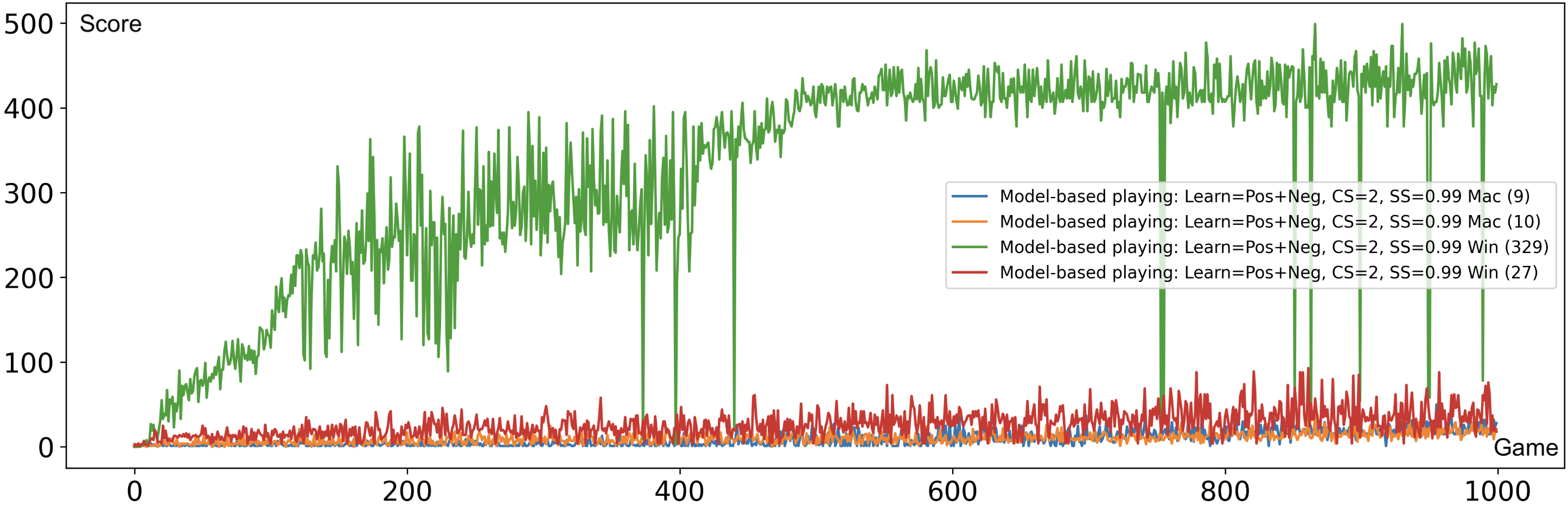}}
    \caption{
      Scores obtained in four different runs on different computers while playing 1000 games with a state similarity threshold of SS=0.99. Horizontal axis - games from 1 to 1000. Vertical axis - scores per game. Plots in different colors correspond to different uncontrolled random seeds. The context size is CS=2. The Win and Mac labels in the legend correspond to the computers on which the respective run was run (Win: MSI Raider, Mac: MacBook Pro). The numbers in parentheses in the legend indicate the average scores.
    }
    \label{fig:round2}
  \end{center}
\end{figure*}

The same framework was further cursory to briefly test the ability to learn from scratch. Testing was conducted using a “Model-based” agent across several different runs with different context size settings (CS=1, CS=2) and state similarity thresholds (SS=0.8, SS=0.9, SS=0.99, SS=0.999), as well as different learning modes (LM=1, LM=2). The initial value of the random number generator was not controlled, as in the previous phase. The evaluation was conducted in batches of 1000 games, with no limit on the number of steps per game. Typical results are shown in \autoref{fig:round2}.

The evaluation showed that achieving high results when training from scratch with CS=1 is impossible, although we observed improved performance with C=1 in the case of the pre-trained model presented in \autoref{sec:phase1}. This is clearly explained by the need to consider the direction of the ball's movement based on at least two consecutive states.

With a context size of two consecutive states (CS=2), the results obtained from the first 1,000 games were inconsistent. Most runs resulted in average scores ranging from 15 to 60 and maximum scores of up to 100, which still exceeded the human average score of 31. However, in some runs, we observed the ability to quickly master gaming skills, so that after several hundred games, the “Model-based” agent reached a score above 400, maintaining it consistently high with rare occasional dips, as shown in \autoref{fig:round2}. Moreover, for such high-performing (“quick learner”) runs, where high scores were achieved within the first 1,000 games, up to 19 additional runs of 1,000 games were run re-using the same model, run by run. In this case, we observed stable performance at an average score above 400, gradually and slightly improving with occasional observation of top 864 scores. For example, \autoref{fig:round2} shows different sets of runs with different random seeds and the same state similarity threshold settings (SS=0.99).

Experiments conducted at this phase with different hyper-parameters confirmed the potential ability of the proposed algorithmic framework to learn from scratch, achieving a stable level of average performance outperforming (under certain random seeds) many known baseline models such as human (31), DQN (168) according to \citet{mnih2013playingatarideepreinforcement}, Rainbow-IQN (176) according to \citet{toromanoff2019deepreinforcementlearningreally}, MARTI-5 (50) according to \citet{pivovarov2025marti5mathematicalmodelself}, and sometimes reaching the highest score of 864 achieved by MuZero \cite{Schrittwieser_2020}.

The experiments confirmed the inappropriateness of using the context size CS=1 and proposed a context similarity (SS) threshold in the range of 0.9–0.99 with a transition utility threshold TU=0, which was used as the initial hyper-parameter setting in the next stage of experiments.

\subsection{Phase 3: Exploring Learning Ability - Systematic}

\begin{figure*}[ht]
  \vskip 0.2in
  \begin{center}
    \centerline{\includegraphics[width=1.7\columnwidth]{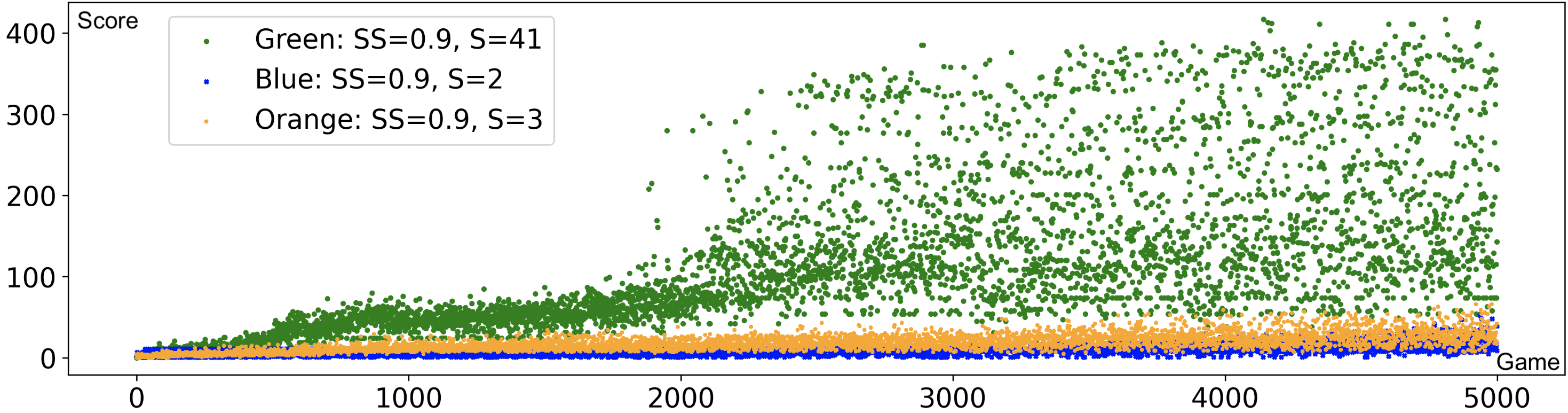}}
    \caption{
      Scores obtained in three different runs, including 5000 games with three different fixed random seeds for the state similarity threshold SS=0.9. Horizontal axis - games from 1 to 5000. Vertical axis - scores per game. Scatter points of different colors correspond to different random seeds S (green – S=41, blue – S=2, orange – S=3). Context size CS=2.
    }
    \label{fig:round3_scatters}
  \end{center}
\end{figure*}

\begin{figure*}[ht]
  \vskip 0.2in
  \begin{center}
    \centerline{\includegraphics[width=1.7\columnwidth]{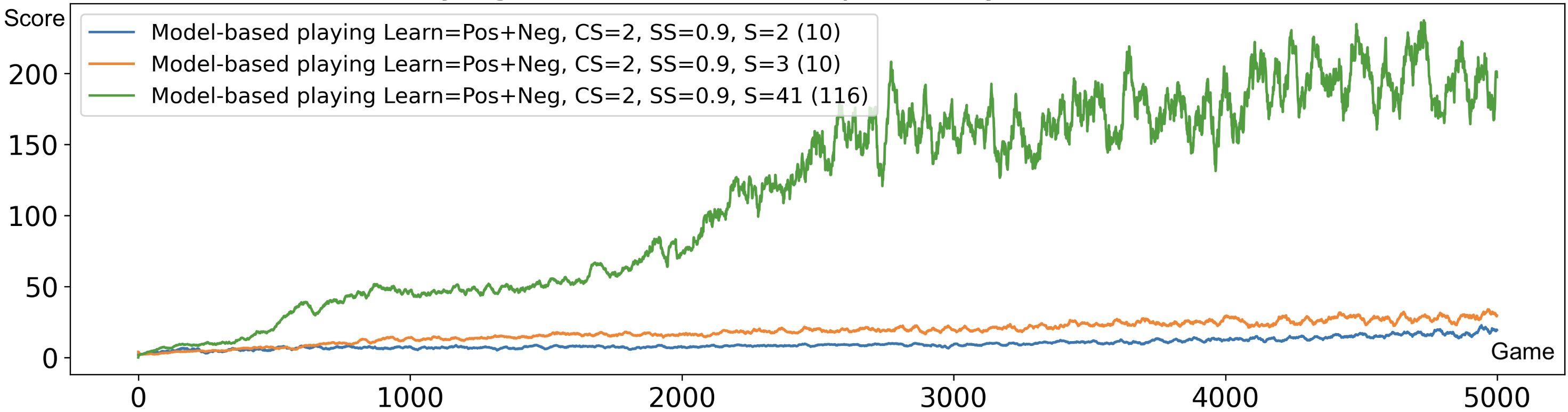}}
    \caption{
      Average scores in a sliding window of 30 games across three different runs, including 5000 games with three different fixed random seeds for the state similarity threshold SS=0.9. Horizontal axis - games from 1 to 5000. Vertical axis - scores per game. Plots in different colors correspond to different random seeds S (green – S=41, blue – S=2, orange – S=3). Context size CS=2. Numbers in parentheses in the legend show average scores across all 5000 games.
    }
    \label{fig:round3_plots}
  \end{center}
\end{figure*}

\begin{figure}[ht]
  \vskip 0.2in
  \begin{center}
    \centerline{\includegraphics[width=0.85\columnwidth]{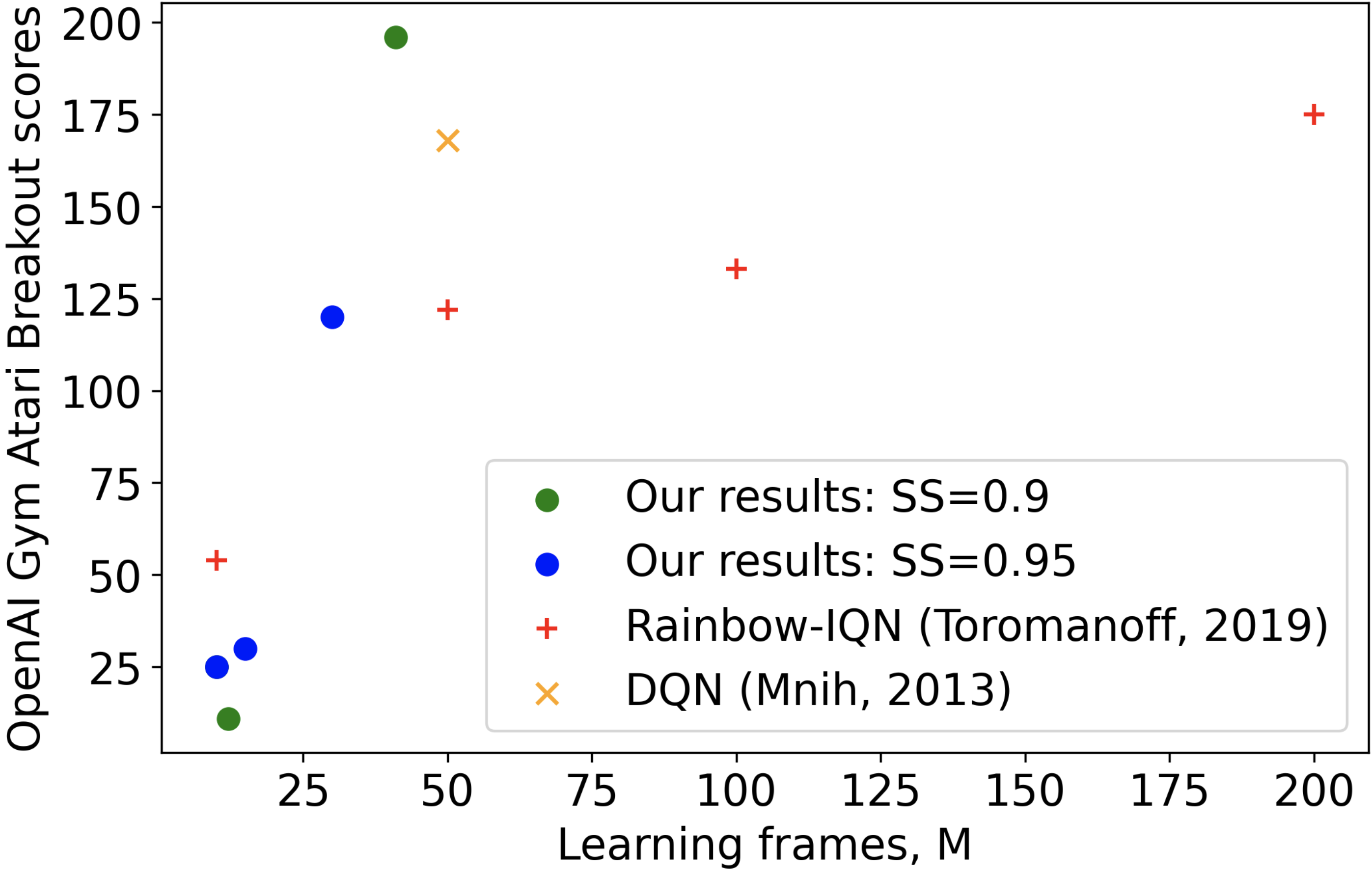}}
    \caption{
      Comparison of the results obtained in different runs of our system with different random seeds (2, 3, 41), and state similarity thresholds (SS=0.9 and SS=0.95) with the results obtained in the works \citet{mnih2013playingatarideepreinforcement} and \citet{toromanoff2019deepreinforcementlearningreally}, depending on the number of frames used for learning.
    }
    \label{fig:comparison}
  \end{center}
\end{figure}

Based on the cursory exploration using uncontrolled random seeds described above, a final phase was to conduct a systematic reproducible study using different fixed random seeds, the results of which are presented in \autoref{fig:round3_scatters}, \autoref{fig:round3_plots}, and \autoref{fig:comparison}.
 
First, 30 different random seed values were evaluated using the default hyper-parameters (LM=2, TU=0, CS=2, TC=1, SC=2, SR=True, SS=0.9, CU=$False$, EA=$False$) according to the results of the previous study. The evaluation was conducted based on 1000 games, and the average scores were collected, with the resulting average score values ranging from 3 to 30.

Based on this, the best performing random seed value of 41 was chosen to search for the optimal values of the state similarity threshold hyper-parameter. Runs were run with all other hyper-parameters set to their default values and SS values of 0.75, 0.8, 0.85, 0.9, 0.95, 0.99, and 0.999. Based on that, SS=0.9 and SS=0.95 were found to be the optimal values based on the average values across all 1000 games (25.8 and 31, respectively). Based on SS=0.9 being the best value in both the previous and present phases of the study, a search was conducted for the other hyper-parameters discussed in the \autoref{hyper-parameters} section. The results confirmed the default hyper-parameter values determined in the previous cursory study, as stated below along with all explored options.

\begin{itemize}
    \item CS=2 (context size, values: 1, 2, and 3) 
    \item LM=2 (learning mode: 0 - no learning, 1 - only from positive feedback, 2 - from both positive and negative feedback) 
    \item SR=True (state reward, values: $True, False$) 
    \item CU=False (counted utility, values: $True, False$) 
    \item EA=False (encode action, values: $True, False$)
    \item SC=2 (state count threshold, values: 1, 2, 3)
    \item SS=0.9 or SS=0.95 (state similarity threshold, values: 0.75, 0.8, 0.85, 0.9, 0.95, 0.99, and 0.999) 
    \item TU=0 (transition utility threshold, values: $None$, 0, 1, 2) 
    \item TC=1 (transition count threshold, values: 1, 2, 3)
\end{itemize}

Based on the distribution of scores for the 30 seeds, 7 representative seeds were selected, and another evaluation was conducted for these 7 seeds using 1000 games with state similarity thresholds of SS=0.9 and SS=0.95, using the remaining default hyper-parameters specified above. Based on the results for these 7 seeds and 2 similarity thresholds, we selected the three seeds (S) that yielded the highest average score (S=41), the lowest average score (S=2), and a score between the highest and lowest (S=3) for further exploration. Using the selected three initial seeds (2, 3, 41), we re-evaluated the learning ability of our system on 5000 games using the same state similarity thresholds of SS=0.9 and SS=0.95. The results are presented in \autoref{fig:round3_scatters} and \autoref{fig:round3_plots}. For the six evaluation runs, we collected the scores of the last 500 games from each run according to \citet{pivovarov2025marti5mathematicalmodelself} and presented the results in \autoref{fig:comparison}.

When attempting to explain the significant differences between the maximum and average results obtained across runs using different random seeds, as shown in \autoref{fig:round2}, \autoref{fig:round3_scatters}, and \autoref{fig:round3_plots}, we found a large variability in the number of steps per game across runs. Therefore, we can hypothesize that shorter games, in terms of the number of steps, may reduce the learning rate estimated based on the number of games. To analyze this, we presented the obtained results based on the number of game frames (equivalent to the number of steps in our setup) used to achieve these results, as shown in  \autoref{fig:comparison}.

\section{Discussion}
\label{discussion}

\subsection{Interpretation and Comparison with Prior Art}

The key observation is that for certain random seeds, we can achieve the same or higher level of learning efficiency for the Atari Breakout game in OpenAI Gym, relative to the number of training frames, compared to known baseline approaches based on DQN \cite{mnih2013playingatarideepreinforcement} and Rainbow-IQN \cite{toromanoff2019deepreinforcementlearningreally} or on the neuro-physiologically plausible design in MARTI-5 \cite{pivovarov2025marti5mathematicalmodelself}. In particular, referring to \autoref{fig:comparison}, we were able to obtain an average score of 120 based on 30 million training frames and an average score of 196 on 41 million frames. At the same time, DQN \cite{mnih2013playingatarideepreinforcement} showed an average score of 168, corresponding to 50 million frames, while Rainbow-IQN \cite{toromanoff2019deepreinforcementlearningreally} achieved scores of 54, 122, 133, and 175 after 10, 50, 100, and 200 million frames, respectively. The only interpretable alternative, MARTI-5 \cite{pivovarov2025marti5mathematicalmodelself}, showed an average score of 50 after training on 170 million training frames. This means that our system can surpass human performance (a score of 31) after a maximum of 30 million frames and achieve a score of over 100 using less than 50 million training frames, while the alternative solutions mentioned above can achieve this level starting from 50 million frames. We attribute the effectiveness of our solution to the “global feedback” principle \cite{10.1007/978-3-030-93758-4_12} used in our system.

This performance was achieved using a model consisting of a transparent state transition graph based on a history of states that does not include hidden or latent variables or states, which is fully interpretable and satisfies both local and global interpretability conditions \cite{bassan2024localvsglobalinterpretability}. Local interpretability means that any action recommended by the framework in a given situation can be traced back to a prior state transition, as well as to the known values of these states and transitions and alternative options in a given situation. All of this can be found in the model, which represents historical transition graphs at a given context depth. Global interpretability implies that transition graphs can be used for process analysis \cite{Koschmider_2024,khan2025advancesprocessoptimizationcomprehensive} to identify meaningful patterns and understand the model's behavior.

\subsection{Limitations of the approach}

Our solution currently lacks the implementation of multi-agency, motivation for exploratory activity and planning features present in \cite{kapturowski2019recurrent,badia2020uplearningdirectedexploration,badia2020agent57outperformingatarihuman,Schrittwieser_2020}, which may explain why our results are below the baseline presented in these studies.

Our experimental methodology might be incomplete, as it was based on the number of games rather than steps. The ability to ensure learning stability regardless of initial random seeds needs to be explored and possibly improved.

The pre-processing required by our current implementation to reduce the state dimensionality and make it game-specific cannot be considered universal and may provide an unfair competitive advantage when compared to other baselines based on raw pixels without manual feature engineering.

\section{Conclusion}
\label{conclusion}

We proposed, implemented, and evaluated an interpretable architecture for experiential or reinforcement learning, based on a model consisting of weighted transition graphs between a series of successor states, with complex weights corresponding to the utility and evidence count of these transitions. The architecture was successfully tested using the OpenAI Gym Atari Breakout environment, showing competitive results against several baselines such as DQN and IQN-based models, and outperforming them in certain conditions when using low-end computing environments.

In future work, we plan to improve the stability of our architecture's learning performance, expand its applicability to various environments, and explore the possibility of redesigning the pre-processing layer for learning the generalized interpretable representation to improve scalability and generalizability. We will also consider extending our architecture to support multi-agency, exploratory activity and planning capacity to achieve higher scores, while continuing to build on our interpretable and resource-efficient approach.

\section*{Impact Statement}

The proposed approach has the potential to enable the development of interpretable solutions using reinforcement learning and experiential learning principles for robust applications in mission-critical and resource-constrained computing environments such as industrial automation, energy, home automation, embedded systems, and edge computing.

Representing a model as a weighted transition graph between interpretable state sequences with state variables corresponding to real-world objects and events allows for its verification, validation, and adjustment. Graph transition weights, assigned based on the utility and evidence counts, can make model analysis more practical and informative. Linking state transition evidence count data to system logs can also make the model fully auditable.

Experimental evaluation of the model on the OpenAI Gym Atari Breakout benchmark confirmed its ability to run in real time on low-end computing infrastructure with competitive performance. Preservation of the evidence counts in the model allows for model compression based on the evidence count thresholds, making the model even more efficient in terms of execution speed and response time.




\bibliography{example_paper}
\bibliographystyle{icml2026}




\end{document}